\documentclass{article}

\usepackage{arxiv}

\usepackage[utf8]{inputenc} 
\usepackage[T1]{fontenc}    
\usepackage{hyperref}       
\usepackage{url}            
\usepackage{booktabs}       
\usepackage{amsfonts}       
\usepackage{nicefrac}       
\usepackage{microtype}      
\usepackage{lipsum}
\usepackage{graphicx}
\usepackage{pifont}
\usepackage{makecell}
\usepackage{colortbl}
\graphicspath{ {./images/} }

\title{Pairwise Spatiotemporal Partial Trajectory Matching for Co-movement Analysis}

\author{
Maria Cardei \\
  University of Virginia\\
  Charlottesville, VA 22903 \\
  \texttt{cbr8ru@virginia.edu} \\
   \And
 Sabit Ahmed \\
  University of Virginia\\
  Charlottesville, VA 22903 \\
  \texttt{bcw3zj@virginia.edu} \\
  \And
 Gretchen Chapman \\
  Carnegie Mellon University\\
  Pittsburgh, PA 15213 \\
\texttt{gchapman@andrew.cmu.edu} \\
    \And
Afsaneh Doryab \\
 University of Virginia\\
  Charlottesville, VA 22903 \\
  \texttt{ad4ks@virginia.edu} \\
}

\begin{document}
\maketitle
\begin{abstract}
Spatiotemporal pairwise movement analysis involves identifying shared geographic-based behaviors between individuals within specific time frames. Traditionally, this task relies on sequence modeling and behavior analysis techniques applied to tabular or video-based data, but these methods often lack interpretability and struggle to capture partial matching. In this paper, we propose a novel method for pairwise spatiotemporal partial trajectory matching that transforms tabular spatiotemporal data into interpretable trajectory images based on specified time windows, allowing for partial trajectory analysis. This approach includes localization of trajectories, checking for spatial overlap, and pairwise matching using a Siamese Neural Network. We evaluate our method on a co-walking classification task, demonstrating its effectiveness in a novel co-behavior identification application. Our model surpasses established methods, achieving an F1-score up to 0.73. Additionally, we explore the method’s utility for pair routine pattern analysis in real-world scenarios, providing insights into the frequency, timing, and duration of shared behaviors. This approach offers a powerful, interpretable framework for spatiotemporal behavior analysis, with potential applications in social behavior research, urban planning, and healthcare.
\end{abstract}


\section{Introduction}





Understanding movement patterns and routine behavior is crucial in a wide range of applications, such as personalized route recommendations,
ride-sharing,
collaborative robotics, and 
social-physical behavior research 
\cite{solomon_analyzing_2021, cui_personalized_2018, sun_neural_2023, zhou_self-supervised_2021}. 
The widespread use of smartphones and wearable devices among individuals yields a vast amount of data that might contain individual, paired, or grouped behavior patterns \cite{banerjee_inferring_2016}.
Studies have demonstrated effective estimation of point-of-interest (POI) or future GPS trajectories and facilitating route recommendations based on historical location data \cite{gambs_show_2010,nascimento_trajectory_2010,endo_predicting_2017,tsiligkaridis_context-aware_2022,zhou_self-supervised_2021,ye_efficient_2024,solomon_analyzing_2021,imai_early_2018,do_places_2014}. Other research efforts have focused on modeling pedestrian movement dynamics using geolocation data \cite{sun_neural_2023,and_modeling_2014,moussaid_walking_2010,zanlungo_potential_2014}. These models can predict pedestrian density and distribution, which can enhance crowd management and improve the design of pedestrian spaces \cite{sun_neural_2023}.  Computer vision models have also been used to identify the physical and social activities of individuals from video clips \cite{ryoo_human_2011,sun_recursive_2020,huang_stgat_2019,10.1007/978-3-030-58545-7_11}. 

One of the major challenges in modeling pairwise or group behavior is determining if two or more people are engaged in a shared activity. For example, many people may walk the same path simultaneously, but only a few walk together. While some existing research has aimed to model complex social and group dynamics of human behavior \cite{zheng_urban_2014,wang_distributed_2020,jeberson_retna_raj_disaster_2010,bond_development_2007}, very little has focused on how to identify co-occurrences of individual, paired or grouped behavior patterns from day-to-day life, which might be essential in various human-centered applications. 
Besides, trajectory matching and forecasting methods cannot provide granular-level views of trajectories, fail to capture time-specific behavioral patterns and routines, and are not interpretable enough for real-world use. Additionally, such methods cannot handle missing, noisy, and sparse geolocation data with incomplete context and may provide erroneous insights \cite{zhou_self-supervised_2021}. While vision-based methods can recognize activities and capture people's interactions through scene understanding, they require continuous monitoring of a person, which is not always possible and is privacy-sensitive. 

We propose a pairwise partial trajectory matching and routine behavior analysis framework to address some of the aforementioned challenges. We evaluate our method in a co-walking detection scenario using data collected from a 5-week pair-walking study involving 126 participants. To our knowledge, our method is the first to classify instances of paired walking activities with notable precision. We extract the partial trajectories of paired walking across different time windows spanning over 24 hours from each day of geolocation data. Each time window is then grouped by several layers, resulting in a sequence of trajectories for different partitioning windows. Next, we preprocess and transform each windowed raw time-series trajectory into image space, resulting in a spatiotemporal object that visualizes trajectories in a given time frame with predefined color gradients. Using a Siamese Neural Network, we detect and classify the partial trajectories of co-walking events in a given time window. In addition to evaluating the performance of the framework in detecting co-walking instances, we demonstrate the potential of the method in extracting paired routing behavior. 
The main contributions of this paper are as follows:

\begin{enumerate}
    \item We propose an interpretable, image-based 
    method to effectively match partial spatiotemporal trajectories of pairwise behavior. We perform an ablation study to assess all the steps of our method to ultimately provide an optimal configuration of our method. 
    \item We evaluate our method by performing experiments to classify co-walking, a novel co-behavior application, and compare performance against established methods. Although computationally slower, it outperforms established methods across classification evaluation metrics.
    \item We demonstrate the capability of our method to detect pair routine patterns and to provide insight into the frequency, timing, and duration of co-behaviors.
\end{enumerate}

\section{Related Work}

\subsection{Trajectory Matching and Destination Forecasting}

Matching movement trajectories of people and early prediction of destinations can accommodate accurate personalized route or travel recommendations, and is crucial in social and robotics navigation, ride-sharing and other service-providing applications \cite{solomon_analyzing_2021, cui_personalized_2018,ye_efficient_2024, wang_understanding_2020}. 
Existing partial trajectory-based predictive systems mainly rely on statistical methods and data-driven approaches. 
Gambs et al. \cite{gambs_show_2010} leveraged the Markov model to predict a person's destination based on the previously visited place information and the associated transition probability distributions. Nascimento et al. \cite{nascimento_trajectory_2010} proposed a two-layer Hidden Markov Model (HMM) to predict pedestrian movement dynamics based on historical trajectories. 
Endo et al. \cite{endo_predicting_2017} represented the sequences of locations as discretized features in a grid space and applied Recurrent Neural Network (RNN) to estimate the transitional probabilities in the next timestamp. The destination probabilities of candidate places were estimated by simulating the movements of objects based on stochastic sampling with an RNN encoder-decoder network.
Solomon et al. \cite{solomon_analyzing_2021} introduced a new method for detecting GPS points where individuals generally reside (e.g., their home, restaurants, working place) and additionally leveraged a Long-Short Term Memory (LSTM) network to forecast possible future locations of travel.
Tsiligkaridis et al. \cite{tsiligkaridis_context-aware_2022} adopted a Transformer-based architecture to predict the destination from partial trajectories and a regression model to estimate the steps required to reach the destination.
Zhou et al. \cite{zhou_self-supervised_2021} argued that data-driven methods, i.e., Markov models, rely on presumed parameters or manually designed features that are not generalizable across different scenarios and may provide misleading interpretations. 
Furthermore, typical machine-learning models may not capture the spatiotemporal nonlinearity from large-scale sparse geolocation data. The authors introduced a Self-supervised Mobility Learning (SML) framework that encodes the geographical and temporal knowledge into contrastive motion representation learning by discriminating the true favorable locations from negative locations, thus maximizing the mutual information between the past observed trajectories and the future mobility.

Imai et al. \cite{imai_early_2018} explain that future destination prediction is one of the most important factors in user behavior prediction. They leverage trajectory tracking and next place prediction (NPP) approach to develop a robust early destination prediction method.
Using collaborative filtering, Cui et al. \cite{cui_personalized_2018} estimated people's travel behaviors from GPS coordinates and developed two personalized travel route recommendation techniques.
Ye and Gombolay \cite{ye_efficient_2024} proposed a new trajectory forecasting and generation framework, Trajectory Control Flow Matching (T-CFM), that transforms trajectory and action data distributions via a learned time-varying vector field. This method has shown efficacy in three different robotics tasks, i.e., adversarial tracking, aircraft trajectory forecasting, and long-horizon planning. 
Most of these studies focus on forecasting the trajectories of individual people and estimating their next places or destinations and the time or steps required to reach them.

Minimal research has been done on matching partial trajectories of two people who might be interacting with each other
or classifying between-people pairwise activities in a constrained-free environment, which can reveal the behavioral dynamics of joint partners.
To this end, we propose a pairwise spatiotemporal trajectory matching system and show it can effectively discriminate the movement trajectories of a pair of individuals.

\subsection{Behavior Analysis From Human Mobility}

Considerable research has been done in analyzing movement and behavioral patterns, which has numerous applications, such as transport management, urban planning, public health and safety, personalized services and marketing, and disaster and emergency management. \cite{wesolowski_connecting_2016,bond_development_2007,jeberson_retna_raj_disaster_2010,mahmood_gps_2006,farrahi_probabilistic_2010,zheng_urban_2014,wang_distributed_2020,zhou_self-supervised_2021,wang_understanding_2020}.
Cuttone et al. \cite{cuttone_understanding_2018} conducted a study to investigate which factors influence the accuracy of predicting the next place location of people.
The results show that predicting the time-bin locations is more accurate than predicting the next places. Furthermore, the resolution of the spatiotemporal data strongly influences the accuracy of the location prediction.
Research efforts by Do et al. \cite{do_places_2014} focused on understanding human mobility patterns from the sequences of places people visit daily and automatically labeling places by leveraging those patterns without requiring any GPS sensor data.
Zhang et al. proposed a data-driven states refinement LSTM (SR-LSTM) network to predict socially aware pedestrian trajectory \cite{9261113}.
Chevance et al. \cite{chevance_characterizing_2021} conducted a study to understand the day-to-day patterns of changes in physical activity behavior and found that fluctuations in walking behavior are associated with sudden irregularities or abrupt behavior changes in subsequent days.
Wang et al. \cite{wang_distributed_2020} presented a large-scale study in Beijing, China, that used taxi GPS data and reported the spatiotemporal characteristics of travel made by the city residents. The travel intensity, duration, and distance on weekdays and weekends were utilized to explore human mobility patterns.


Pedestrian flow models can be leveraged to predict the walking dynamics of people in crowded places \cite{sun_neural_2023}. 
One study suggests that pedestrians moving in groups, also referred to as group dynamics, might affect the walking behavior of individual pedestrians \cite{and_modeling_2014}.
Many studies aim to investigate the statistical properties of group dynamics \cite{sun_neural_2023,moussaid_walking_2010}. Moussaïd et al. \cite{moussaid_walking_2010} found that the size of a group influences the walking speed and formation of a social group. 
Another research proposed a model for the dynamics of the relative position of two pedestrians in a group \cite{zanlungo_potential_2014}. They also reported that a small group of pedestrians usually forms distinct shapes, such as V, when three people are in that group.
According to this model, pedestrian walking behavior is influenced by the individual's goal and desired walking speed, physical boundaries or obstacles like walls or fences, and attractive effects of surrounding objects or agents in the environment. Furthermore, the grouped walking behavior of pedestrians can be considered an important factor influencing the individual pedestrian's walking patterns. Sun et al. \cite{sun_neural_2023} derived a neural network model incorporating variables that influence a pedestrian's movement behavior in social groups to model such walking dynamics.

While a variety of studies aim to obtain individual or group behavior insights from small—or large-scale mobility data, there is a lack of co-behavioral pattern analysis, which could potentially identify the time-dependent behavior or activities of single or multiple people. Our proposed pairwise trajectory matching method can transform the raw mobility data into visually interpretable trajectories, thus revealing co-occurrences of routine behavior across one or more people.

\subsection{Computer-vision based Activity Recognition}
Instead of GPS-based trajectory detection methods, several studies use computer-vision models to recognize mobility or activity patterns. M. S. Ryoo \cite{ryoo_human_2011} developed an early activity recognition system that analyzes streaming videos of ongoing activities. Ehsanpour et al. \cite{10.1007/978-3-030-58545-7_11} introduced a framework for social activity recognition that predicts individual activities, social interactions, group dynamics, and the overall social activities of groups using spatiotemporal representations derived from video clips. An Inflated 3D ConvNet (I3D) architecture was used, where ImageNet pre-trained convolution kernels were efficiently expanded to 3D for the spatio-temporal representation.
Sun et al. \cite{sun_recursive_2020} proposed a novel graph representation and generation framework named Recursive Social Graph for social behavior modeling from video scenes. This framework can extract latent pedestrian relationships and integrate human social behaviors in dynamic scenes for predictive tasks.
Huang et al. \cite{huang_stgat_2019} combined a graph attention network (GAT) with LSTM to encode the spatial and temporal interactions among pedestrians and forecast future trajectories of pedestrians.
By understanding the walking behavior of nearby pedestrians, Kayukawa et al. \cite{kayukawa_guiding_2020} developed a camera-based guiding system to enable blind people to walk in public spaces.

Existing vision-based models involve complex scene understanding, region of interest detection, trajectory simulation, and activity recognition from streaming or recorded videos. However, video-capturing devices or surveillance systems are not necessarily available or accessible in remote areas or even social gathering places and often fail to capture complex relationships or interactions among people from the image frames. 
In contrast, our method relies solely on a GPS sensor, a common feature in most smartphones and wearable devices such as smartwatches and rings. Our proposed method converts the extracted trajectories into image data, allowing us to localize and identify paired activities from these images.

\section{Pairwise Spatiotemporal Partial Trajectory Matching}


\begin{figure}
    \centering
\includegraphics[width=1\linewidth]{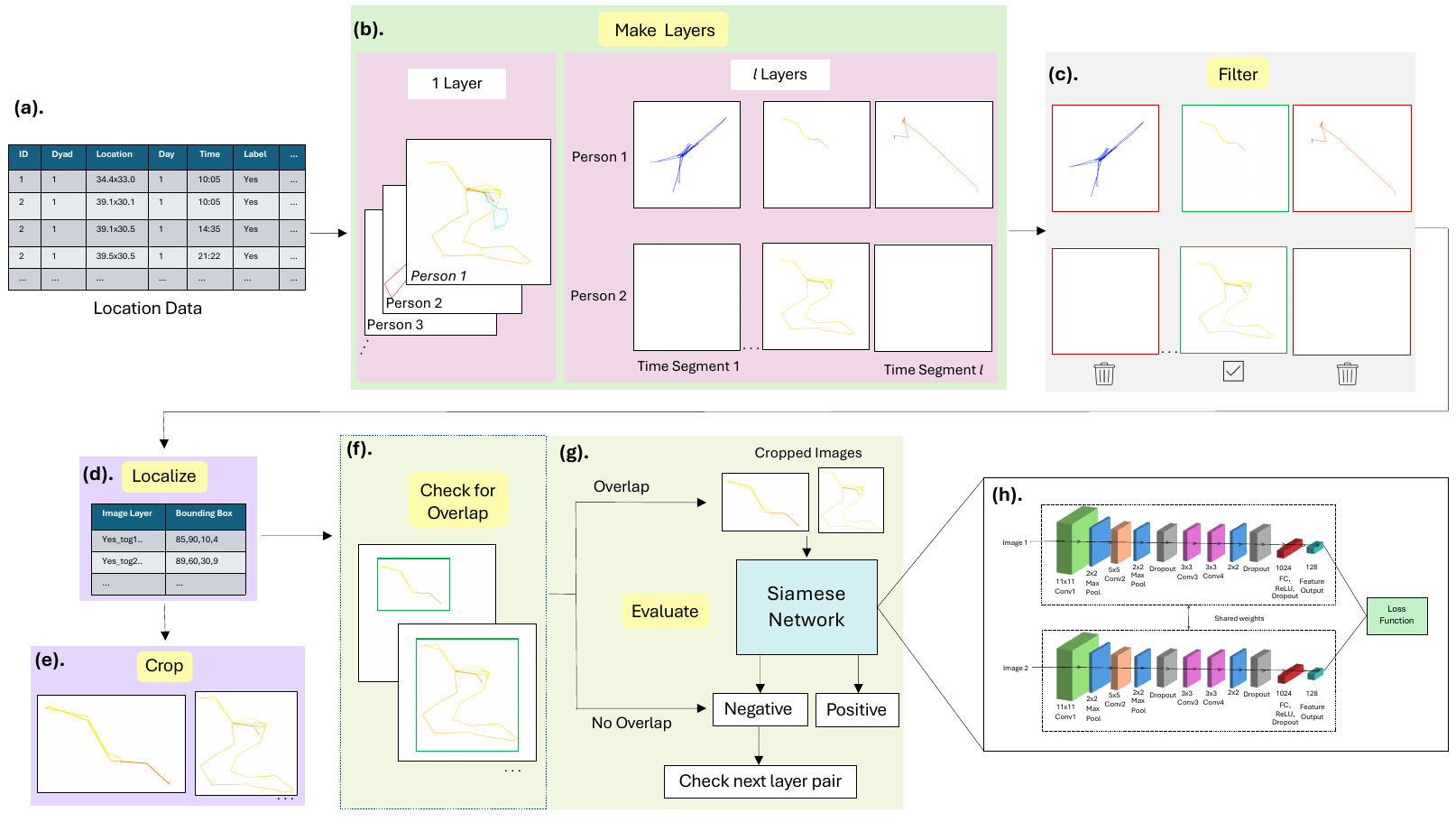}
    \caption{Proposed method pipeline, where $l$ refers to the number of layers. The steps are labeled (a)-(h). The Siamese Network in (h) is the SigNet model, adapted from \cite{signet}.}
    \label{fig:pipeline}
\end{figure}

We propose a framework to 
identify pairwise spatiotemporal partial trajectory matching. While this method can be applied beyond pairs and GPS data, this paper specifically examines its effectiveness in detecting pairwise mobility using GPS location. The framework involves generating image layers, filtering out poor-quality images, trajectory localization, image cropping, checking spatial overlap, and Siamese Neural Network evaluation (Fig. \ref{fig:pipeline}). This method turns raw, tabular location data into a colorful, interpretable visualization of spatiotemporal data. The image layers enable the division of data into time intervals, allowing for partial trajectory classification. This not only helps with the classification of trajectory matching but also allows for improved data interpretation for routine pattern analysis, which will be discussed further. The following sections describe the components of the framework. 

\label{section:method}
\subsection{Transform Tabular Location Data Into Image Layers}
The first step uses spatiotemporal data collected from devices like wearables or smartphones, where each timestamp is linked to a specific geographic location, typically represented by latitude and longitude coordinates. This data should include location information for two individuals at each timestamp, allowing for a detailed and interpretable analysis of whether trajectories match during specific periods. 

We generate layered visual representations of trajectories by dividing the spatiotemporal data into distinct time intervals or layers (Fig. \ref{fig:pipeline}b). Each layer corresponds to a specific portion of the total analysis period, enabling us to observe detailed movement patterns over time. This allows us to perform partial trajectory matching, as the entirety of the data is divided into time intervals. The user can define the total time duration and choose the number of layers based on the available data and the desired level of detail. For instance, to analyze movement over a 24-hour period, the total time can be divided into 24 layers, each representing one hour. In this case, layer one would display movement between 12:00 am and 1:00 am, layer two between 1:00 am and 2:00 am, and so on. The number of layers defines how many images will be generated and the level of detail in each image. Location data is mapped as points in space and connected with lines to create a visual movement trajectory. Time data is normalized within each time interval and represented using a custom color map, where gradients show the progression from the start to the end of the time period. For a pair of individuals, an image is generated for each person for each time interval, with all images sharing the same canvas size. The canvas is resized according to the combined latitude and longitude ranges of the trajectory pairs during that interval, facilitating easy spatial comparisons. 

When comparing the images from a pair, the location of the trajectories in the images represents their spatial locations, while the color of the trajectory indicates the time of movement within that interval. 
An image is created for each individual, covering each specified time window, with all images utilizing the same canvas size.
Layers are generated by providing the pair's location data. This way, the canvas coordinates will represent the latitude and longitude information considering both pairs' trajectories for the given time interval and will be scaled together. For example, when comparing two images from a pair of individuals to assess co-walking, the physical location of the trajectories indicates the physical location coordinates and the color of the trajectory indicates the time. This technique results in a highly interpretable visualization of spatiotemporal data.

\subsection{Filter Out Poor-Quality Image Layers}
Some image layers may have minimal or no trajectories, especially as the number of layers increases to show details on a smaller scale, such as over a few hours, minutes, or even seconds. Since each layer represents a shorter time interval, there may be fewer timestamps with movement data. This is particularly true when co-behavior is less likely, such as at night. To address this and poor-quality data, we develop a step to filter out low-quality image layers (Fig. \ref{fig:pipeline}c). Our method analyzes each image layer by counting the number of colored pixels and calculating statistical metrics, such as the minimum, average, maximum, and interquartile values. Using these metrics, along with empirically determined values, we establish a pixel threshold that balances removing noise while retaining enough data for reliable analysis. The threshold value is chosen based on the available data and its characteristics. Layers that fall below this threshold are excluded. To ensure consistency when comparing image pairs of two individuals who may have engaged in co-behaviors, we discard a layer for both individuals whenever it is removed for one of them.

\subsection{Localize Trajectories and Crop Images}
We localize the trajectories to aid the final evaluation (Fig. \ref{fig:pipeline}d). The trajectory can be anywhere within the canvas, thus we implement OpenCV's Selective Search algorithm to generate bounding boxes around each trajectory \cite{opencv_segmentation}. We save the coordinates of the bounding box encompassing the entire trajectory. Each image layer is then cropped to focus solely on the trajectory, minimizing the surrounding white space (Fig. \ref{fig:pipeline}e).

\subsection{Check Spatial Overlap and Evaluate}
Trajectories may look alike while being spatially located in different areas within their images. Therefore, we compare a pair of image layers to check that their respective areas overlap within the canvas.
Starting with the first layer pair and continuing through the rest, we check for overlap between the two images (Fig. \ref{fig:pipeline}f). 
We use the saved bounding box values for both images to check for any overlap between them, determining whether the boxes occupy similar spatial areas. If there is no overlap between the bounding boxes, we classify the layer pair as negative. This indicates that the individuals' trajectories do not align during that time segment, prompting us to proceed to the next layer. This approach ensures that if the pair was not in the same physical space during a time segment, they are definitively classified as not having engaged in co-behaviors together. However, if there is overlap, the pair's cropped image layers are passed to a Siamese Neural Network, which classifies the layer as positive or negative (Fig. \ref{fig:pipeline}g). If no overlap is found, or if the Siamese Network classifies the layer as negative, the next layer pair is considered. This process continues until a layer pair is classified as positive, indicating that the trajectories match, or until all layers of that instance have been checked. This method not only determines whether a pair engaged in behaviors together but also identifies the specific time interval based on the layer number when the trajectories align.

\subsubsection{Siamese Neural Network} A Siamese Network (Fig. \ref{fig:pipeline}h) is used to evaluate pairs of images by learning a similarity function. It takes two images as input and outputs a probability indicating how similar the images are. This architecture is ideal for our classification task because it only requires a single trajectory image to calculate a similarity score with other trajectories and can classify based on a similarity score threshold. Given its ability to perform one-shot classifications and the minimal data required, this architecture is well-suited for determining co-behavior. Since trajectories can vary across time intervals, days, and individuals, the flexibility of the Siamese Network makes it an ideal fit for our problem.

A Siamese Network contains two identical sub-networks, often composed of Convolutional Neural Networks (CNNs). 
Both inner sub-networks share the same parameters and weights, with synchronized parameter updates. 
Each inner sub-network processes an input image to generate a feature representation that captures the unique characteristics of that image. 
During training, the output features from each of the sub-networks are compared with a Contrastive Loss function to minimize the distance between feature vectors for inputs from the same class while maximizing the separation for inputs from different classes. The mathematical formulation of Contrastive Loss is as follows:
\begin{equation}
    L(s_1, s_2, y) = \alpha(1 - y)D_w^2 + \beta y \max(0, m - D_w)^2  ,
\end{equation}
where $s_1$ and $s_2$ represent two images, $y$ is a binary indicator function, $\alpha$ and $\beta$ are constants, and $m$ is the margin
\cite{signet}. \( D_w \) is the Euclidean distance between the embedded feature vectors \( f(s_1; w_1) \) and \( f(s_2; w_2) \), where \( f \) is an embedding function mapping an image to a vector space using a CNN with learned weights $w_1$ and $w_2$ for each underlying layer in the network \cite{signet}. 
Euclidean distance quantifies the proximity of features in the learned space, enabling the Siamese Network to classify input pairs as similar or distinct based on spatial closeness.
With both sub-networks embedding images into a unified space, a final layer calculates the Euclidean Distance between the embeddings at the time of inference. A set threshold is then applied to classify the pair as either similar (positive) or dissimilar (negative) based on the computed distance.

\section{Experimental Evaluation}
To evaluate our proposed method for pairwise spatiotemporal partial trajectory matching, we address the novel problem of classifying co-walking scenarios—an area not previously explored. Specifically, we utilize spatiotemporal data, comprising time and location, from pairs of individuals to match their partial trajectories. This enables us to determine whether they have walked together, providing new insights into movement patterns and behavioral tendencies.

\subsection{Data Collection and Processing} \label{section:dataset}

We utilize a private dataset from a university 
to classify co-walking. The dataset was collected during a 5-week study that aimed to test the effects of incentives on encouraging co-physical activity and reducing loneliness. Participants were recruited in pairs and were encouraged to walk with their partners every day during the study. Each participant would earn one dollar for meeting their steps goal and one additional dollar for walking at least a mile with a walking partner. Participants' locations were tracked using a FitBit and the smartphone application AWARE \cite{aware_paper}. Data was collected when movement was detected. The study involved 126 college students and their friends 
and included 79 females (62.7\%) and 41 males (32.5\%). 
Participants were instructed to complete a daily survey at the end of each day, in which they self-reported whether they walked with their partner that day.

We took multiple steps to clean and prepare the raw data for modeling and analysis. 
Since the dataset was collected during the start of the COVID-19 pandemic, the protocol allowed for virtual co-walking, i.e., walking at the same time but at different routes and locations while talking on the phone. Given that our goal is to model and detect physical co-walks, we removed all virtual pairs. We also adjusted the timestamp to local time to ensure the timing of co-walks are correctly captured. 
If one individual in a pair had no data on a given day, that day's data for themselves and their partner were removed. We also removed any data for which the ground-truth label (i.e., self-report indicating whether the pair co-walked) was missing. If the self-reports from both individuals in a pair did not match for a particular day—meaning one person reported that they walked together while the other claimed they did not—the data for that day was labeled as not having walked together. It is important to note that daily self-reports only provided information about whether or not the pairs walked but not when and where they walked together. As such, we needed to assess trajectories throughout the day and perform partial trajectory matching to find the time window and trajectory the pair walked together. 
We aimed to preserve the authenticity of real walking scenarios and demonstrate that our method performs well even when some data points are missing. As such, we did not interpolate missing data.

We utilize location (latitude, longitude), time, pair information, and daily survey responses to assess our proposed framework to classify whether pairs co-walked on a given day. Based on participants who completed the full study and were not excluded in our pre-processing, the resulting dataset consisted of 94 unique individuals over 38 different days. This resulted in 1,472 instances, with each instance representing one person's data for a single day. Of these, 507 instances involved co-walking, while 965 did not. Participant pairs averaged 10 co-walking days (Min = 1, Max = 26, Std = 6.998). 

\subsection{Image Layer Generation and Modeling} \label{Experiment_setup}

\begin{figure}
    \centering
\includegraphics[width=1\linewidth]{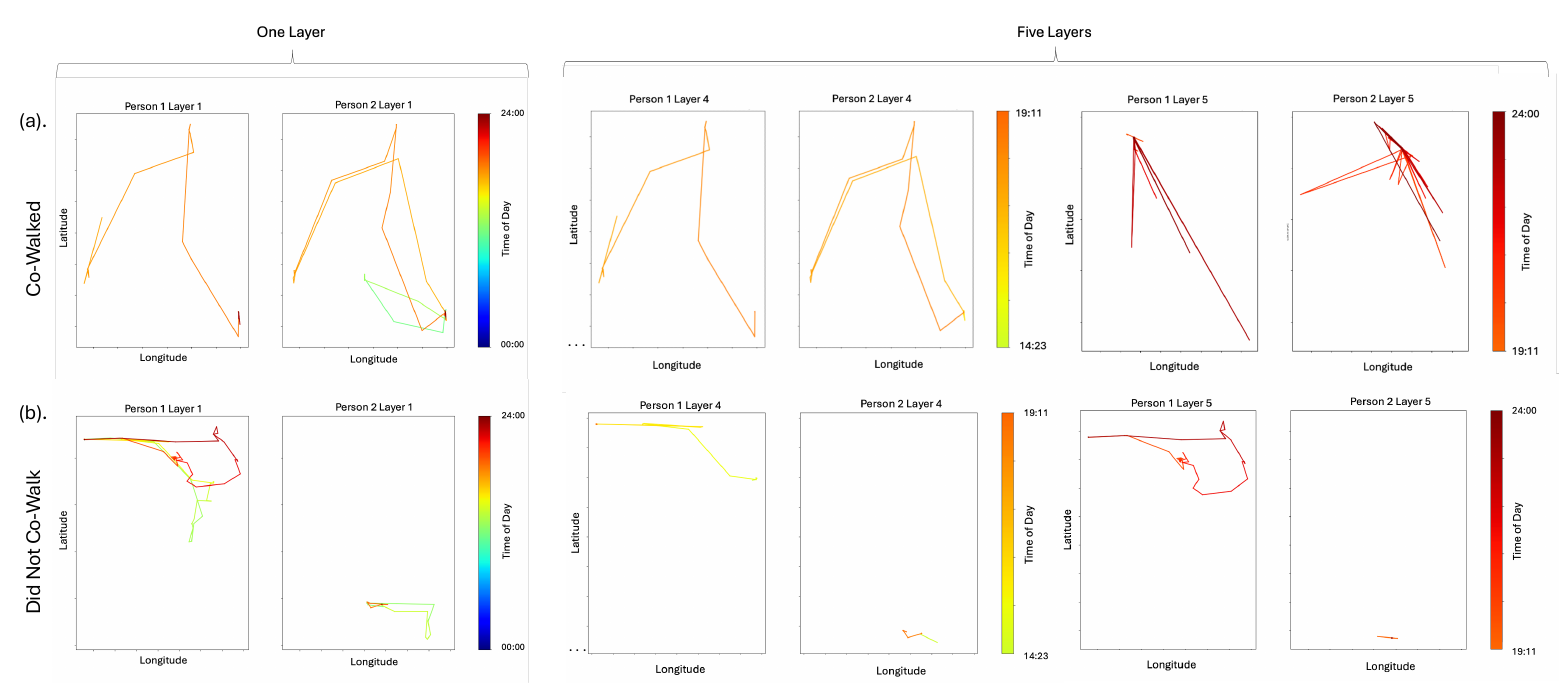}
    \caption{Detailed view comparing a pair that co-walked on a day (row a) and a pair that did not co-walk on a day (row b). The day is split into one layer images as well as five layer images, where layers 4 and 5 are shown as examples. Both individuals' locations are scaled together within the same latitude and longitude values to fit on the canvas per layer. With five layers, each layer allows for a more detailed view of the locations for each time segment, essentially 'zooming in' to each location visited (e.g., the latest time interval of 19:11-24:00 visualized in red is zoomed in for the five layer scenario as compared to the red trajectory in the one layer scenario).}
    \label{fig:layer_views}
\end{figure}

\subsubsection{Generating Layers}

We experimented with different sets of layers for generating images that represent an individual's daily trajectories. 
To assess the impact of varying the number of layers, we generate \textbf{1, 5, 24, and 48 layers} for each individual and their partner across all days of data. One image is generated for each layer. Figure \ref{fig:layer_views} demonstrates an example of layers created for a pair that co-walked on a given day and for a pair that did not. This figure highlights the differences in trajectory detail achieved for 1 versus 5 total time segments (layers). Capturing one layer per day enables us to visualize an individual's daily movements in a single image, which contains a wealth of information. By dividing this data into five layers, we can reveal significantly more detail. We also experiment with using 24 layers to represent one hour's worth of data in each image. To gain a clearer understanding of an individual's daily movement, we create 48 image layers each day, providing access to data in half-hour increments.
We assume the time of co-walking can vary and may be as brief as a half-hour or even less. Consequently, with 48 layers, we achieve substantial insight into individual patterns. 
Additionally, comparing the impact of different layer counts helps us understand how varying levels of trajectory detail in each image influence classification performance.

\subsubsection{Filtering} As we generate more layers from a day's data, we visualize trajectories over smaller time windows. During some periods, an individual's movement may be minimal or absent. To enhance the accuracy of our analysis, we filter out these images, as the absence of a trajectory indicates that no co-walking could have occurred during those times. We calculate statistical metrics for the images, including the minimum, average, and maximum number of colored pixels on the white canvas, along with the $25^{th}$, $50^{th}$, and $75^{th}$ percentile values. This analysis is conducted for the 1-layer, 5-layer, 24-layer, and 48-layer sets of images to capture metrics specific to each set. These metrics provide insight into the data distribution, helping us retain relevant data while avoiding discarding too much or too little. Based on empirical testing and the analysis of these metrics, we apply a \textbf{filtering threshold at the $\mathbf{25^{th}}$ percentile for the 1-layer images and at the $\mathbf{50^{th}}$ percentile for the 5-layer, 24-layer, and 48-layer images.}

\subsubsection{Siamese Neural Network}
For our classification task, we use an existing Siamese Network architecture called SigNet \cite{signet}. SigNet was originally trained on signature images, taking in two images of signatures and identifying whether the input image is genuine or forged. We chose SigNet because the signature images resemble the trajectory visualizations we create and thus, can extract similar features i.e., centroids, shapes, or edges, etc. SigNet has demonstrated impressive accuracy, achieving up to 100\% on detecting forged signatures within a dataset and up to 94.82\% accuracy across datasets, with most results ranging between 50-77\% accuracy \cite{signet}. We aim to test how well this model performs on our similar but distinct GPS trajectory dataset. We experiment with both the \textbf{pre-trained SigNet model (PT model) and fine-tuned SigNet models (FT models)}. We leverage the pre-trained SigNet model to infer the images generated from our preprocessed trajectory data. To obtain more accurate classifications, we then adopt a transfer learning approach. We trained the SigNet model with our generated images, thus making one trained Siamese network for each set of image layers.
This allows us to compare the effects of using a pre-trained model versus adapting the model with task-specific data. We implement 5-fold cross-validation to ensure robust evaluation and generalization of our models. We train the networks for 50 epochs and evaluate the model performance aligning with the cross-validation steps. This reduces the risk of overfitting and ensures results are not dependent on a specific train-test split. Further details of SigNet can be found in Figure \ref{fig:pipeline}h and in the paper \cite{signet}.

\subsection{Benchmark Methods}
\label{subsec: sota} Current established techniques in trajectory classification and identification are predominantly deep learning-based. They are highly effective at capturing both spatial and temporal dependencies in trajectory data. However, to our knowledge, there is no existing intuitive, visualization-based classification technique like the one we propose. Therefore, we employ an LSTM (Long Short-Term Memory) and a GRU (Gated Recurrent Unit) to classify co-walking based on raw spatiotemporal data. LSTMs and GRUs are widely used and well-regarded benchmarks in sequence modeling, particularly in temporal data contexts. They are effective baselines for tasks requiring temporal dependency modeling, and comparing against these allows us to establish a performance baseline within the familiar recurrent framework. Though potentially applicable, more sophisticated architectures, such as Transformers, focus on temporal attention and sequence alignment rather than spatial interpretability.


\textbf{LSTMs} are Recurrent Neural Networks (RNN) designed to handle sequential data by capturing long-term dependencies and temporal patterns. They are particularly useful for co-walking classification because they can model the temporal dependencies between consecutive location points. 
We incorporate a two-layer LSTM with dropout regularization. It utilizes Xavier and orthogonal initialization methods for its weights to promote stable learning and maintain variance. We also implement an early stopping mechanism to avoid over-fitting. 


\textbf{GRUs} are another variant of RNNs designed to handle sequential data. They are similar to LSTMs, however GRUs have fewer parameters and use gating mechanisms to control the flow of information, making them computationally more efficient while still effectively capturing temporal dependencies in sequences. We employ a GRU with two layers and dropout regularization. During training, an early stopping mechanism is integrated to monitor validation loss and stop training if no progress is made. 

We implement 5-fold Cross-Validation in our training of these models to continue to ensure robust evaluation. We train both models for 50 epochs. The results from these models are then compared with those from our proposed method, which uses trajectory visualizations and image classification. The raw latitude and longitude values and the pair's identifying information and dates are utilized. To address data imbalance, we compare the lengths of trajectory pairs and retain only those with matching lengths between individuals. This process minimizes redundant trajectory pairs that might not accurately represent the positive class.

\subsection{Performance Measures}

We perform classification with the pre-trained SigNet Siamese network and the fine-tuned SigNet models, where we train on our own image data sets. We report \textbf{precision, recall, accuracy,} and \textbf{F1 score}. These are classic metrics that provide a comprehensive view of classification performance: precision indicates how many of the predicted positives were correct, recall measures how many actual positives were accurately identified, accuracy reflects the proportion of correctly classified instances, and the F1 score balances precision and recall, providing insight into the model's effectiveness in handling imbalanced data.
We also measure \textbf{Matthews Correlation Coefficient (MCC)}, a robust metric for binary classification. MCC considers true and false positives and negatives, giving a more balanced view of model performance, especially when class distributions are skewed. MCC values range from -1 (perfect misclassification) to +1 (perfect classification), with 0 indicating no predictive power. Finally, we evaluate \textbf{execution time}, which indicates the computational efficiency of each method. The times listed result from utilizing the NVIDIA General Purpose GPUs. 
Lower execution times imply faster model inference, which can be crucial for scalability. 

\subsection{Results}\label{results}
We generated 1, 5, 24, and 48 image layer sets, implementing the proposed method described in Section \ref{section:method}. The proposed method includes the pre-trained (PT) SigNet Siamese model, trained on signature image data, and the fine-tuned (FT) SigNet Siamese model, trained on our co-walking images. In this section, we compare the performance of our proposed method with benchmark techniques, discuss the interesting effects observed in an ablation study, and evaluate the execution time.

\subsubsection{Comparing to Benchmark Methods}
 As shown in Table \ref{tab:sota_results}, 
our PT Siamese Network outperforms the GRU across all evaluation metrics. However, some layered image sets performed on par with, or worse than the LSTM. Notably, the 24 and 48 image layer sets consistently outperformed the established methods, however the 1 layer set performed worse than the LSTM, while the 5 layer set performed about evenly with a lower precision and recall score, but higher accuracy and MCC values. 

The FT models demonstrated improved performance compared to the PT models, with lower execution times across all layers. 
The FT Siamese models outperform LSTM and GRU across evaluation metrics, with a particularly notable improvement in the MCC score (0.36 for 24 layers, compared to 0.16 and 0.06, respectively). This indicates that numeric values cannot capture the representative features of co-walking trajectories, while our image-based approach provides a more robust and accurate classification. However, our method requires considerably more execution time than LSTM and GRU models. This trade-off indicates that while our proposed method, which utilizes image layer generation, outperforms time-series models that depend on tabular data, it also leads to greater computational cost. Therefore, our approach is best suited for applications where high accuracy is prioritized over speed, and where computational resources are not a constraint. Our method is better suited for situations that require an interpretable representation of data and the ability to identify the timing and frequency of co-behavior, which is less feasible with existing methods.


\begin{table}[t]
    \centering
        \caption{Average weighted results of the proposed method compared to benchmark methods, which are trained on time-series spatiotemporal data. The proposed method includes the pre-trained (PT) SigNet Siamese model, trained on signature image data, and the fine-tuned (FT) SigNet Siamese model, trained on our co-walking images implementing 5-fold cross-validation. We report results for all layers tested, utilizing our entire proposed method.
        }
    \begin{tabular}{cccccccccc}
        \toprule
         & Method & {\makecell{No. Layers}} & Precision & Recall & Accuracy & F1-Score & MCC & {\makecell{Execution\\Time (s)}} \\
        \midrule
         &  PT Siamese & 1 & 0.59 & 0.59 & 0.58 & 0.58 & 0.16 & $3.7 \times 10^{-1}$ \\
          & & 5 & 0.61 & 0.59 & 0.60 & 0.58 & 0.21 & $4.3 \times 10^{-1}$ \\
&& 24  & 0.68 & 0.64 & 0.65 & 0.63 & 0.32 & $5.0 \times 10^{-1}$ \\
           Proposed & & 48 & 0.66 & 0.62 & 0.64 & 0.62 & 0.29 & $5.8 \times 10^{-1}$ \\
        \cmidrule(lr){2-9}
& FT Siamese &1  & 0.65 & 0.61 & 0.62 & 0.60 & 0.25 & $2.6 \times 10^{-1}$ \\
          & & 5 & 0.67 & 0.58 & 0.60 & 0.55 & 0.25 & $2.3 \times 10^{-1}$ \\
&& 24 & \textbf{0.71} & 0.64 & \textbf{0.67} & 0.63 & \textbf{0.36} & $2.4 \times 10^{-1}$ \\
           && 48 & 0.70 & \textbf{0.65} & \textbf{0.67} & \textbf{0.64} & 0.35 & $2.3 \times 10^{-1}$ \\
           \hline
        Benchmark  & LSTM & - & 0.62 & 0.61 & 0.57 & 0.58 & 0.16 & $1.4 \times 10^{-4}$ \\
         & GRU  & - & 0.53 & 0.59 & 0.54 & 0.54 & 0.06 & \textbf{\boldmath$8.1 \times 10^{-5}$}
\\
        \bottomrule
    \end{tabular}
    \label{tab:sota_results}
\end{table}

\begin{table}[ht]
    \small
    \centering
    \caption{Average weighted results of each ablation study step utilizing the SigNet model both pre-trained (PT) on signature images and fine-tuned (FT). The FT models are trained on our co-walking images and weighted average metrics are reported, after performing 5-fold cross validation. The best metrics are in bold for each column.}
    \begin{tabular}{@{} l l | l|l | l|l | l|l | l|l | l|l | l|l@{}}
    \toprule
    & \makecell{No. \\ Layers} & \multicolumn{2}{c|}{Precision} & \multicolumn{2}{c|}{Recall} & \multicolumn{2}{c|}{Accuracy} & \multicolumn{2}{c|}{F1 Score} & \multicolumn{2}{c|}{MCC} & \multicolumn{2}{c}{\makecell{Execution\\Time (s)}} \\ 
    \cmidrule(lr){3-4} \cmidrule(lr){5-6} \cmidrule(lr){7-8} \cmidrule(lr){9-10} \cmidrule(lr){11-12} \cmidrule(lr){13-14}
    & & PT & FT & PT & FT & PT & FT & PT & FT & PT & FT & PT & FT \\ 
    \midrule
           & 1 &  0.52 & 0.57 & 0.43 & 0.42 & 0.47 & 0.49 & 0.43 & 0.38 & -0.06 & -0.02 & \textbf{0.13} & 0.23 \\
           & 5 &  0.53 & 0.12 & 0.34 & 0.34 & 0.50 & 0.50 & 0.18 & 0.18 & -0.01 & 0.00 & 0.17 & 0.16 \\
Baseline   & 24 & 0.11 & 0.26 & 0.33 & 0.35 & 0.50 & 0.50 & 0.17 & 0.18 & 0.00 & 0.02 & 0.20 & 0.35 \\
           & 48 & 0.11 & 0.12 & 0.33 & 0.34 & 0.50 & 0.50 & 0.17 & 0.18 & 0.00 & -0.03 & 0.24 & \textbf{0.13} \\
\midrule

           & 1 & 0.70 & 0.75 & \textbf{0.72} & 0.71 & 0.65 & 0.72 & 0.70 & 0.71 & 0.32 & 0.42 & 0.33 & 0.27 \\
           & 5 & 0.72 & 0.75 & 0.71 & 0.71 & 0.69 & 0.72 & 0.71 & 0.71 & 0.36 & 0.42 & 0.34 & 0.21 \\

Overlap    & 24 & \textbf{0.74} & 0.75 & \textbf{0.72} & 0.71 & \textbf{0.71} & 0.72 & \textbf{0.73} & 0.71 & \textbf{0.41} & 0.42 & 0.60 & 0.31 \\
           & 48 & 0.73 & \textbf{0.76} & \textbf{0.72} & 0.72 & \textbf{0.71} & \textbf{0.74} & 0.72 & \textbf{0.73} & 0.40 & \textbf{0.45} & 0.62 & 0.60 \\
\midrule

           & 1 & 0.67 & 0.75 & 0.68 & 0.70 & 0.61 & 0.72 & 0.67 & 0.71 & 0.24 & 0.41 & 0.27 & 0.25 \\
           & 5 & 0.71 & 0.72 & 0.70 & 0.68 & 0.68 & 0.68 & 0.71 & 0.68 & 0.36 & 0.34 & 0.32 & 0.22 \\
Overlap \& Crop & 24 & \textbf{0.74} & 0.70 & \textbf{0.72} & 0.70 & \textbf{0.71} & 0.67 & 0.72 & 0.70 & \textbf{0.41} & 0.34 & 0.57 & 0.36 \\
           & 48 & 0.72 & 0.74 & 0.70 & \textbf{0.73} & 0.70 & 0.71 & 0.71 & 0.73 & 0.38 & 0.41 & 0.79 & 0.84 \\
\midrule

           & 1 & 0.64 & 0.70 & 0.64 & 0.64 & 0.62 & 0.66 & 0.64 & 0.63 & 0.26 & 0.34 & 0.41 & 0.33 \\
           & 5 & 0.62 & 0.65 & 0.60 & 0.61 & 0.60 & 0.62 & 0.58 & 0.59 & 0.22 & 0.25 & 0.58 & 0.34 \\
Overlap \& Filter & 24 & 0.68 & 0.69 & 0.64 & 0.62 & 0.66 & 0.63 & 0.63 & 0.59 & 0.32 & 0.30 & 0.46 & 0.43 \\
           & 48 & 0.68 & 0.61 & 0.64 & 0.61 & 0.66 & 0.57 & 0.64 & 0.56 & 0.32 & 0.17 & 0.77 & 0.72 \\
\midrule

           & 1 & 0.59 & 0.65 & 0.59 & 0.61 & 0.58 & 0.62 & 0.58 & 0.60 & 0.16 & 0.25 & 0.37 & 0.26 \\
          & 5 & 0.61 & 0.67 & 0.59 & 0.58 & 0.60 & 0.60 & 0.58 & 0.55 & 0.21 & 0.25 & 0.43 & 0.23 \\
Entire Method & 24 & 0.68 & 0.71 & 0.64 & 0.64 & 0.65 & 0.67 & 0.63 & 0.63 & 0.32 & 0.36 & 0.50 & 0.24 \\
           & 48 & 0.66 & 0.70 & 0.62 & 0.65 & 0.64 & 0.67 & 0.62 & 0.64 & 0.29 & 0.35 & 0.58 & 0.23 \\
\bottomrule
    \end{tabular}%
    \label{tab:avg_results_all}
\end{table}







        

\subsubsection{Ablation Study}
While our proposed method involves all steps described in Section \ref{section:method}, we conducted an ablation study to assess the contribution of each component's contribution and validate the necessity to train the Siamese network on our data. Results for the pre-trained (PT) SignNet Siamese model and the fine-tuned (FT) models trained on our co-walking images are shown in Table \ref{tab:avg_results_all}.
For each, we first ran the proposed method, stripping away the filtering, localization, cropping, and overlap verification steps. Then, we added each of the steps iteratively. 'Baseline' refers to the initial model that generates layer images from location data without any added steps (Fig. \ref{fig:pipeline}b), and using these images as input to the Siamese Network for evaluation (Fig. \ref{fig:pipeline}g). 'Overlap' involves generating layer images from location data, localizing trajectories (Fig. \ref{fig:pipeline}d), checking for spatial overlap within the images, and evaluating results (Fig. \ref{fig:pipeline}f, \ref{fig:pipeline}g). 'Overlap \& Crop' involves generating image layers, localizing, and cropping images (Fig. \ref{fig:pipeline}e), checking for overlap, and then evaluating them. 'Entire Method' refers to the addition of the filtering step (Fig. \ref{fig:pipeline}c) alongside the previously mentioned steps, resulting in the complete proposed pipeline.

In the following sections, we discuss the impact of each step on classification performance and execution time. The execution times reported include the overall time in seconds, including generating images, the respective steps for the method, and the classification by the Siamese Network.

\paragraph{Impact of Number of Layers} We tested both the PT and FT Siamese models at each stage of the ablation study. We evaluated image sets with 1, 5, 24, and 48 layers to examine how the number of image layers generated from tabular location time-series data impacts the results. For the \textbf{PT Siamese model}, in general, all ablation steps (except for the Baseline) achieved their best performance with 24- and 48-layer image sets. For the Baseline method, the 1-layer images showed the highest recall (0.43) and F1 score (0.43), along with the shortest execution time. As additional steps were added to the proposed method, the 24-layer image sets consistently performed best overall, achieving the highest precision, recall, accuracy, F1 score, and MCC across all steps. The only exception was in the Overlap \& Filter method, where the 48-layer image set produced a slightly higher F1 score (0.64 vs. 0.63). Although the 24-layer image sets performed comparably to the 48-layer sets in most cases, they consistently required less execution time, making them more efficient.

For the \textbf{FT Siamese models} trained on our co-walking data, results varied more across layers. For Baseline, the 1-layer image set performed best, resulting in a higher precision, recall, and F1 score. For Overlap, results were very similar across all layer image sets, with 48 layer images performing best with a precision of 0.76, recall of 0.72, accuracy of 0.74, F1 score of 0.73, and MCC of 0.45. For Overlap \& Crop, 1 and 48-layer images performed best, while for Overlap \& Filter, 1-layer images outperformed the rest. For the Entire Method, the 24 and 48-layer image sets performed best. 
When there are fewer layers, each layer contains more information, leading to a broader view of the images. This makes it difficult to distinguish smaller trajectories. Additionally, with fewer layers, some trajectories may overlap if an individual remains stationary or repeatedly follows the same path. However, generating more layers can be more computationally expensive.

\paragraph{Impact of Spatial Overlap Check}
We compared the Baseline setting to the Overlap check. Across both the PT and FT models, results indicate that performance metrics drastically improved for all image layer sets in the Overlap check setting compared to the Baseline. In particular, for 24 and 48 image layer sets, the metrics improved the most, with 24 layer images performing best at this stage for the PT model with a precision of 0.74, recall of 0.72, accuracy of 0.71, F1 score of 0.73, and MCC value of 0.41. The 48-layer image set performed best for the FT models, with a precision of 0.76, recall of 0.72, accuracy of 0.74, F1 score of 0.73, and MCC of 0.45. 

\paragraph{Impact of Cropped Images as Siamese Network Input}
When comparing the results for Overlap to the results of Overlap \& Crop for both the PT and FT models, results generally do not improve. Each performance metric stays the same or decreases when the cropped images are fed into the Siamese network. The 24-layer image set with cropping performs almost identically in the PT model case, yielding the best-performing combination with the PT model since the overall execution time is lower.

\paragraph{Impact of Filtering Out Poor Images}
Comparing the results for Overlap \& Filter to prior methods for both the PT and FT models shows that performance generally worsens when poor-quality images are removed before the following steps. Each performance metric stays the same or decreases compared to simple Overlap and Overlap \& Crop. However, the performance is still better than that of the baseline method. This indicates that filtering out images does not always improve the quality of the Siamese network classification.

\paragraph{Classification of Co-Walking}
\begin{figure}
    \centering
\includegraphics[width=1\linewidth]{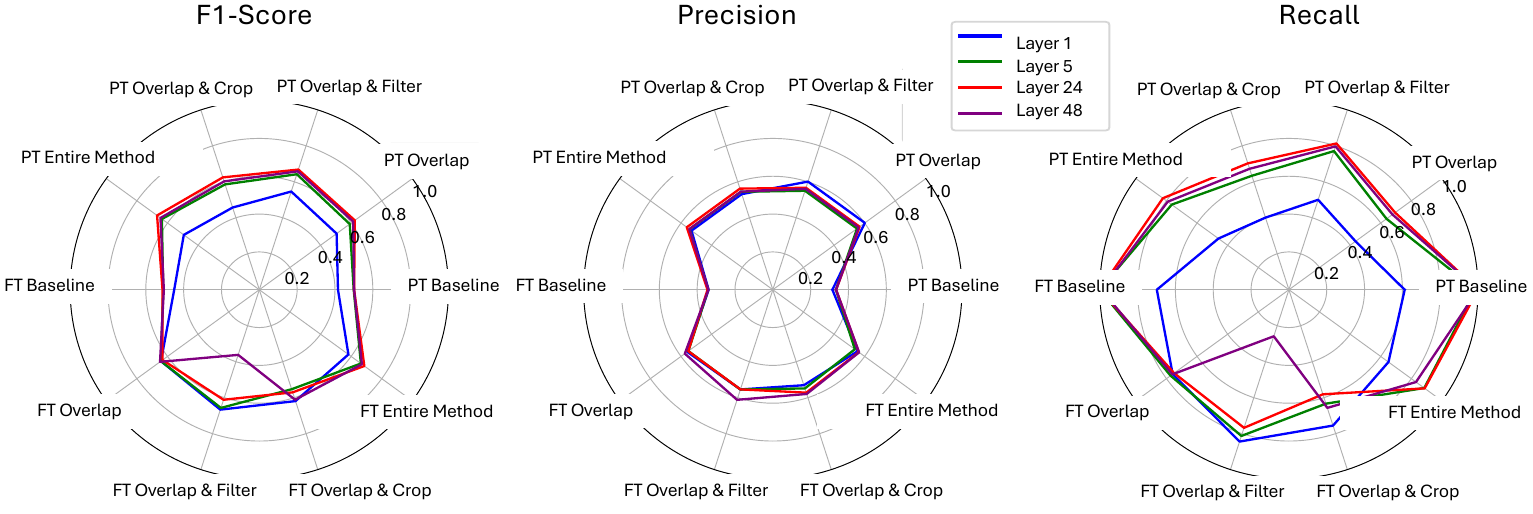}
    \caption{Positive class results (F1-score, precision, and recall) from co-walking classification. Each section of the radar charts aligns with each step of the Ablation Study, with $PT$ indicating the Siamese model is pre-trained on signature data, while $FT$ indicates we trained the model on our image trajectory data.}
    \label{fig:pos_class_radar}
\end{figure}

Figure \ref{fig:pos_class_radar} shows a new angle of the Ablation Study to investigate the F1-score, precision, and recall of the co-walking class classification. Visually, results indicate that the overall metrics are better for 5, 24, and 48-layer image sets. Despite achieving high recall scores, the Baseline models were found to be less successful in terms of precision and F1-score. This suggests that the Baseline models may classify non-co-walking instances as co-walking events, leading to false positives. Additionally, adding a spatial overlap check before classification significantly improved performance over the Baseline experiments. Although some false positive predictions remain, the higher F1-scores and precision achieved by the overlap checking methods indicate they are more effective at accurately identifying true co-walking events. Further steps, such as cropping and filtering, have negligible effects on performance, resembling overall results in Table \ref{tab:avg_results_all}.


\paragraph{Optimal Configuration}
Based on the results, our method performed optimally with a 24-layer image set (PT model case) and a 48-layer image set (FT model case) when including the localization and spatial overlap check but without the filtering or cropping steps. Therefore, our empirical results allow us to propose these as the most effective configurations for our method, as shown in Figure \ref{fig:updated_pipeline} within the Appendix. 

\subsubsection{Computational Cost of Generating Images}

\begin{table}[ht]
    \centering
        \caption{Execution time in seconds for generating each image layer set from time-series location data. The time to generate each individual image layer, as well as the total time for all images, is listed. As the number of layers increases, it takes less time to generate each image; however, it takes more total time to generate all images. This is because images in image sets with fewer layers each contain more information than layers generated from fewer data points.}
    \begin{tabular}{lcccc}
        \toprule
        \makecell{No. Layers} & \makecell{No. Images} & \makecell{Time per Image (s)} & \makecell{Total Time (s)} \\
        \midrule
        1 & 1,388 & 0.05972 & 82.89 \\
        5 & 6,940 & 0.00280 & 97.04 \\
        24 & 33,312 & 0.00041 & 329.66 \\
        48 & 66,624 & 0.00025 & 794.70 \\
        \bottomrule
    \end{tabular}
    \label{tab:make_layers_time}
\end{table}

Table \ref{tab:make_layers_time} depicts the time required to generate images for each layer set and the total number of images produced. Since each layer is generated from a single day of location time-series data when data is divided into a single layer, the entire day’s data is captured in that one image. As the number of layers increases from 1 to 48, the data is split into smaller time intervals, resulting in more images per day. Consequently, higher layer counts yield more images but with less data per image, making each image quicker to generate. As the number of layers increases, the time required to generate each image decreases approximately tenfold with each increase in layer count. However, the total generation time increases as the number of images grows, reflecting the additional processing required.

\begin{figure}
    \centering
\includegraphics[width=1\linewidth]{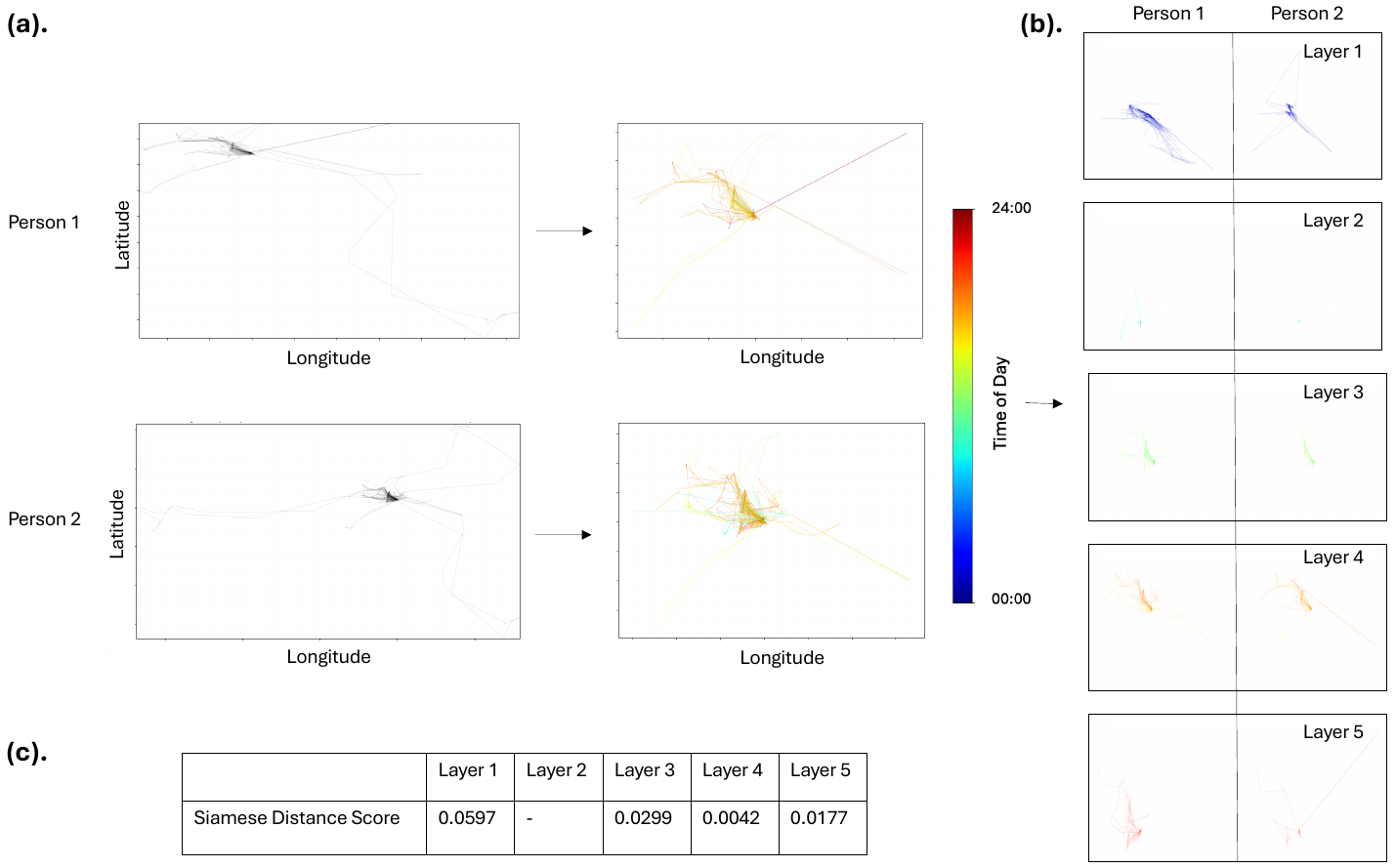}
    \caption{(a) Routine behavior patterns for two individuals in a pair across all days' of data (Person 1: 23 days, Person 2: 26 days). Utilizing clustering, routine patterns are identified and color-coded by time of day. (b) Five-layer images are generated for each individual from all days of data. Each layer represents a different time window of the day, progressing along the custom gradient. (c) When these layers are passed through a Siamese Network, a distance score is obtained for each layer, indicating the similarity between layers and likely co-walking intervals based on routine behavior. In this case, the pair most likely co-walked during the time interval of Layer 4 since it has the smallest distance, followed by Layer 5. Layer 2 images lacked sufficient data for a reliable distance measure.}
    \label{fig:analysis_siamese}
\end{figure}

\subsection{Pair Routine Pattern Analysis}

Beyond identifying co-behavior instances, e.g., co-walking, we explore the method's ability to identify routine patterns in pair behavior. For example, we want to analyze how often, what time, how long, and on what paths a pair goes for walks. 
Figure \ref{fig:analysis_siamese} illustrates how our method can be used for this analysis. We start by generating trajectory plots for all days of data, with overlapping trajectories indicating repeated movement patterns. Using DBSCAN \cite{DBSCAN}, we filter out non-routine paths to retain only the consistent trajectories that represent routine behaviors. Figure \ref{fig:analysis_siamese}a shows this process for two individuals in a pair, with routine trajectories color-coded to represent the time of day, providing an hourly-level view of patterns across many days.

Next, we generate image layers from multiple days of location data to capture finer details of routine behavior. Figure \ref{fig:analysis_siamese}b displays five image layers for the two individuals, shown side-by-side, with each layer representing a different time interval. Visually, we observe similarities between the pair's patterns within each layer, suggesting potential co-walking times. To quantify this, we employ one of the Siamese models trained on our data (FT models used in Section \ref{Experiment_setup}) to assess distance values between individuals for each image layer. The FT models were trained on daily images created from 1, 5, 24, and 48-layer image sets. Here, we use the highest-performing model (the 5th fold model trained on 48-layer images with Overlap only) to evaluate distance values for each time interval. Figure \ref{fig:analysis_siamese}c shows an example, highlighting the intervals during which the pair most likely co-walked, with Layer 4 (2-7pm local time) resulting in the smallest distance value, followed by Layers 5, 3, and 1. Layer 2 was excluded from analysis due to insufficient trajectory data, as we found that sparse data could yield nearly blank images with misleadingly small distance values despite the absence of actual co-walking.

\section{Discussion}

In this work, we proposed a pipeline for pairwise spatiotemporal partial trajectory matching and demonstrated its utility for analyzing pair routine movement patterns. Our approach transforms raw time-series or tabular location data into layered, interpretable image visualizations, capturing both geographic and temporal aspects essential for improved classification performance. This visual format enables the direct recognition of complex spatiotemporal patterns, surpassing established sequence models like LSTM and GRU, which process data only as sequential numerical inputs. By using a Siamese Network architecture, our model effectively identifies subtle similarities and differences between paired trajectories, allowing it to detect partial matching with greater sensitivity. This approach enhances our model's ability to distinguish routine co-behavior with higher accuracy than traditional tabular-data methods. While our method requires additional execution time for image generation and processing, this cost is offset by enhanced classification accuracy and interpretability. For applications like co-walking classification, where precise identification of shared movement is critical, this trade-off is justified, especially when spatiotemporal insights are prioritized over processing speed.

Although methods such as LSTM and GRU models are relevant in temporal modeling, they inherently lack the spatial interpretability that our image-based approach provides.
Our method has demonstrated strong performance even when applied to highly imbalanced datasets with missing data, overcoming limitations that hindered the established text-based methods with the same conditions. Moreover, by integrating clustering techniques, our pipeline supports a detailed analysis of routine movement patterns beyond partial trajectory matching. This method can identify the frequency, time, and duration of coordinated behaviors. Such insights are valuable for applications in social behavior research, healthcare, and urban planning, where understanding routine interactions and changes in movement behavior are essential.

A key finding from our work is that some of the proposed additional steps led to poorer performance in pairwise partial trajectory matching. Our initial pipeline involved generating layered images from raw location data, filtering out poor-quality images, localizing trajectories, cropping the images, checking for spatial overlap, and then classifying with a Siamese Network. However, our ablation study revealed that the Siamese Network achieved better results when we simplified the process to include only image layer generation, localization, spatial overlap verification, and classification. Filtering and cropping were found to be unnecessary for accurate pairwise partial trajectory matching, making this an important insight for researchers working on similar tasks. Through further experimentation with different numbers of image layers, we concluded that sets with 24 and 48 layers performed best. This is likely because these configurations allowed us to capture finer temporal details, such as hourly and half-hourly patterns for our co-walking experiment, which align well with typical co-behavior durations for this case. Since co-walking events can span only a few minutes to hours, these layer configurations effectively highlighted the relevant time intervals.

We also evaluated the performance of a SigNet model pre-trained on signature images (PT model) and fine-tuned it on our own data (FT models). The FT models consistently achieved faster execution times, likely due to being trained on data similar to our target task, allowing them to classify images more efficiently. Additionally, the FT models showed superior evaluation metrics, outperforming the PT model. This demonstrates the value of fine-tuning on task-specific data, though the PT model’s strong performance highlights the effectiveness of transfer learning with similar data types. Overall, these findings validate the effectiveness of our proposed method and underscore the potential of transfer learning in this context.

\subsection{Limitations \& Future Work}
The key limitations are threefold: the method has a relatively high execution time compared to other methods, there can be dataset limitations, and it has a sensitivity to parameter choice. We also discuss privacy-related concerns associated with our method.

Based on the observed execution times from our experimental analysis, the proposed method incurs a higher computational cost for classifying trajectories as compared to that of raw, tabular location data with established methods. This is due to the additional complexity of generating image layers from raw data and the computational demands of utilizing a Siamese Network. However, since co-behavior classification is typically an offline task, this may not present a significant limitation. We envision this approach being generalizable to other applications, such as identifying similar movement patterns among individuals to facilitate carpool assignments. For real-time or online applications, future work will need to focus on optimizing and accelerating the process.

The dataset used in this study presented several challenges. First, it was a small dataset with class imbalance, which complicated accurate classification and limited the generalizability of results. Additionally, the classification relied on individuals to self-report accurately about whether they co-walked, introducing potential biases and inaccuracies. The dataset also contained substantial missing data and inconsistencies, which further impacted the robustness of the analysis. For future work, addressing these limitations will involve sourcing or creating larger, more balanced walking datasets and potentially incorporating automated co-walking detection methods to reduce reliance on self-reporting. Improving data quality and consistency will be essential for enhancing model performance and reliability.

The method also exhibits sensitivity to parameter choices, which were largely determined through empirical analysis on a specific dataset (e.g., selecting 24 and 48 layers for optimal co-walking classification performance). These choices were tailored to the unique characteristics of the data, such as the frequency and duration of co-walking events. It would be essential to apply our technique to various datasets with other pairwise behaviors, such as co-commuting or collaborative work, to evaluate the method's generalizability. Exploring alternative architectures that are less sensitive to specific configurations may also enhance the robustness of the model. Finally, incorporating more automated data quality checks and handling missing data more systematically will improve model performance and make the approach more versatile for broader applications.

While our pairwise spatiotemporal trajectory matching method offers powerful insights into shared movement patterns, it also raises important privacy considerations. Tracking and analyzing partial trajectories involves processing sensitive spatiotemporal data, which could inadvertently reveal personal routines, relationships, or affiliations. Even when data is anonymized, there is a risk of re-identification, as unique movement patterns or frequent locations can often be traced back to individuals. Additionally, analyzing pairwise behaviors may expose social or professional associations that individuals might prefer to keep private. To address these concerns, it is essential to incorporate privacy-preserving techniques, such as data anonymization, differential privacy, and aggregation, reducing the potential for identifying individuals while enabling meaningful analysis. Ensuring transparency and consent is also critical; individuals should be informed about how their data is used, particularly in applications involving social behaviors or shared routines. By implementing these privacy safeguards, our approach can responsibly support applications while respecting individuals' privacy and data security.

\section{Conclusion}
This paper presents a novel pipeline for pairwise partial spatiotemporal trajectory matching, identifying the optimal steps that make our approach highly data-driven. To evaluate our method, we performed an experiment to classify co-walking. Our method achieved an accuracy of 0.74, an F1 score of 0.73, and an MCC value of 0.45, significantly outperforming benchmark methods. Additionally, we demonstrate the method's potential for pair routine pattern analysis. This provides researchers with a valuable tool to uncover insights into the frequency, timing, and duration of co-behaviors, enabling applications in areas such as social behavior research, health monitoring, and urban planning. Future work will focus on optimizing computational efficiency, exploring adaptive parameter selection to enhance generalizability, and validating the method on diverse datasets to expand its utility across domains.

\bibliographystyle{unsrt}  
\bibliography{template}  






\newpage
\appendix
\section{Appendix}
\begin{figure}
    \centering
\includegraphics[width=0.8\linewidth]{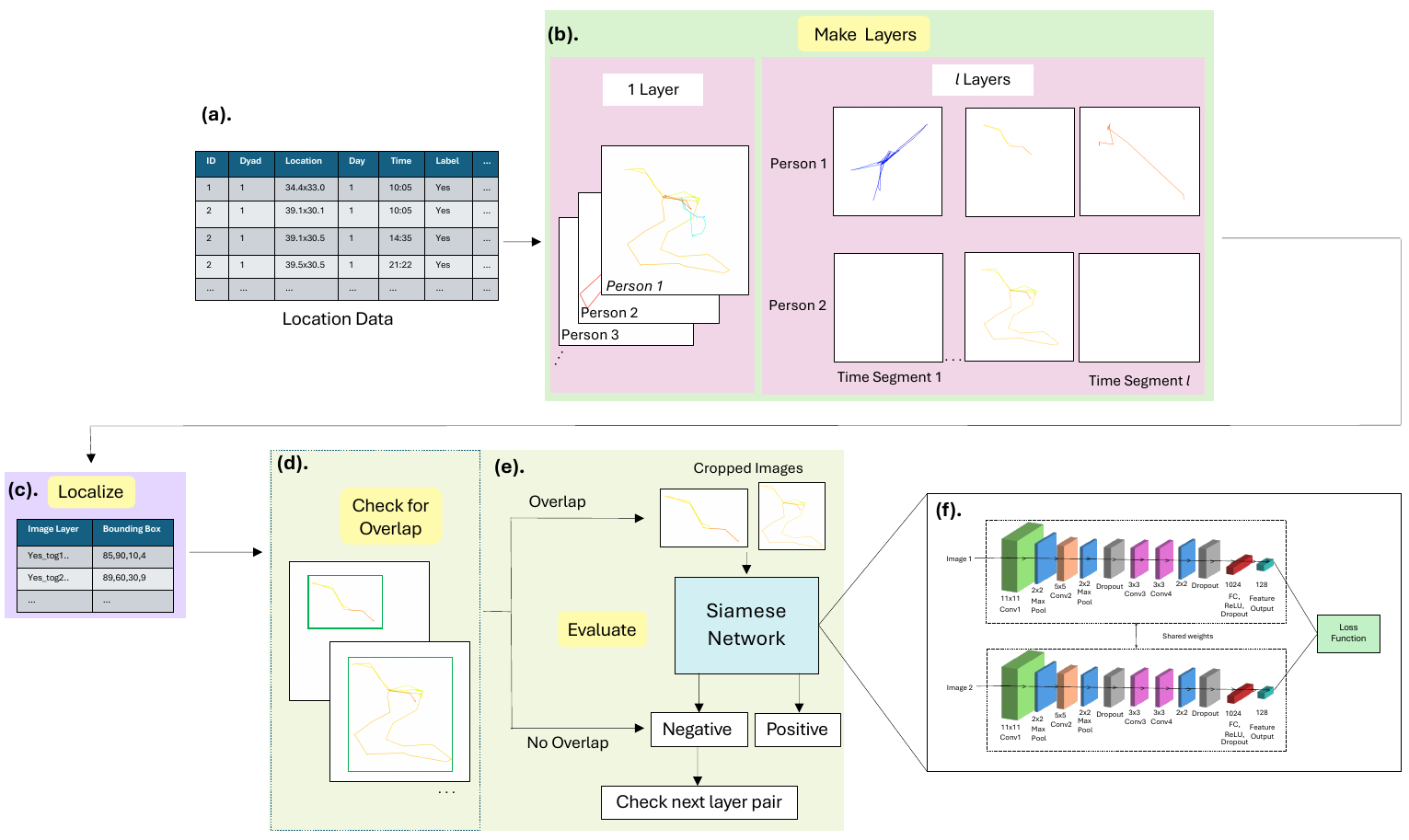}
    \caption{Updated proposed method pipeline, where $l$ refers to the number of layers. The steps are labeled (a)-(f). The Siamese Network in (f) is the SigNet model, adapted from \cite{signet}. This pipeline reflects the optimized configuration of our initial method based on empirical results.} \label{fig:updated_pipeline}
\end{figure}

Figure \ref{fig:updated_pipeline} represents our updated proposed method based on our Ablation Study findings.

\end{document}